\documentclass[10pt,twocolumn,letterpaper]{article}

\usepackage[pagenumbers]{cvpr} 
     
\usepackage{graphicx}
\usepackage{amsmath}
\usepackage{amssymb}
\usepackage{booktabs}
\usepackage{array}    
\usepackage[accsupp]{axessibility}
\usepackage{pifont}
\usepackage{amsmath}
\newcommand{\cmark}{\ding{51}}%
\newcommand{\xmark}{\ding{55}}%

\usepackage{ulem}
\usepackage[pagebackref,breaklinks,colorlinks]{hyperref}

\usepackage[capitalize]{cleveref}
\crefname{section}{Sec.}{Secs.}
\Crefname{section}{Section}{Sections}
\Crefname{table}{Table}{Tables}
\crefname{table}{Tab.}{Tabs.}

\def\eg{\textit{e.g.}}

\def\etc{\textit{etc.}}
\def\confName{CVPR}
\def\confYear{2022}

\begin{document}

\author{Fan Yan\textsuperscript{1}\thanks{The first two authors contributed equally to this work.}
\quad
Ming Nie\textsuperscript{1}$^*$
\quad
Xinyue Cai\textsuperscript{2} 
\quad
Jianhua Han\textsuperscript{2}
\quad
Hang Xu\textsuperscript{2}
\quad
Zhen Yang\textsuperscript{2} \\
\quad
Chaoqiang Ye\textsuperscript{2} 
\quad
Yanwei Fu\textsuperscript{1} 
\quad
Michael Bi Mi\textsuperscript{2} 
\quad
Li Zhang\textsuperscript{1}\thanks{Li Zhang (lizhangfd@fudan.edu.cn) is the corresponding author at School of Data Science and Shanghai Key Laboratory of Intelligent Information Processing, Fudan University} \\
\textsuperscript{1}School of Data Science, Fudan University
\quad
\textsuperscript{2}Huawei Noah’s Ark Lab
\quad
\\
\url{https://once-3dlanes.github.io}
}

\title{ONCE-3DLanes: Building Monocular 3D Lane Detection}
\maketitle

\begin{abstract}
We present~\textit{ONCE-3DLanes}, a real-world autonomous driving dataset with lane layout annotation in 3D space.
Conventional 2D lane detection from a monocular image yields poor performance of following planning and control tasks in autonomous driving due to the case of uneven road.
Predicting the 3D lane layout is thus necessary and enables effective and safe driving.
However, existing 3D lane detection datasets are either unpublished or synthesized from a simulated environment, severely hampering the development of this field.
In this paper, we take steps towards addressing these issues.
By exploiting the explicit relationship between point clouds and image pixels, a dataset annotation pipeline is designed to automatically generate high-quality 3D lane locations from 2D lane annotations in 211K road scenes.
In addition, we present an extrinsic-free, anchor-free method, called \textit{SALAD}, regressing the 3D coordinates of lanes in image view without converting the feature map into the bird's-eye view (BEV). 
To facilitate future research on 3D lane detection, we benchmark the dataset and provide a novel evaluation metric,
performing extensive experiments of both existing approaches and our proposed method.

The aim of our work is to revive the interest of 3D lane detection in a real-world scenario.
We believe our work can lead to the expected and unexpected innovations in both academia and industry.

\end{abstract}

\section{Introduction}
\label{sec:intro}

The perception of lane structure is one of the most fundamental and safety-critical tasks in autonomous driving system.
It is developed with the desired purpose of preventing accidents, reducing emissions and improving the traffic efficiency~\cite{CRAYTON2017245}.
It serves as a key roll of many applications, such as lane keeping, high-definition (HD) map modeling ~\cite{homayounfar2020hierarchical}, trajectory planning \etc~In light of its importance, there has been a recent surge of interest in monocular 3D lane detection~\cite{2020Genlanenet,2021DualATTENTION,20193Dlanenet,20203Dlanenet+,2020semi-local3dlane}.
However, existing 3D lane detection datasets are either unpublished, or synthesized in a simulated environment due to the difficulty of data acquisition and high labor costs of annotation.
With the only synthesized data, the model inevitably lacks generalization ability in real-word scenarios.
Although benefiting from the development of domain adaptation method~\cite{garnett2020synthetic}, it still cannot completely alleviate the domain gap.

\begin{figure}[t]
\centering 
\includegraphics[width = 8cm]{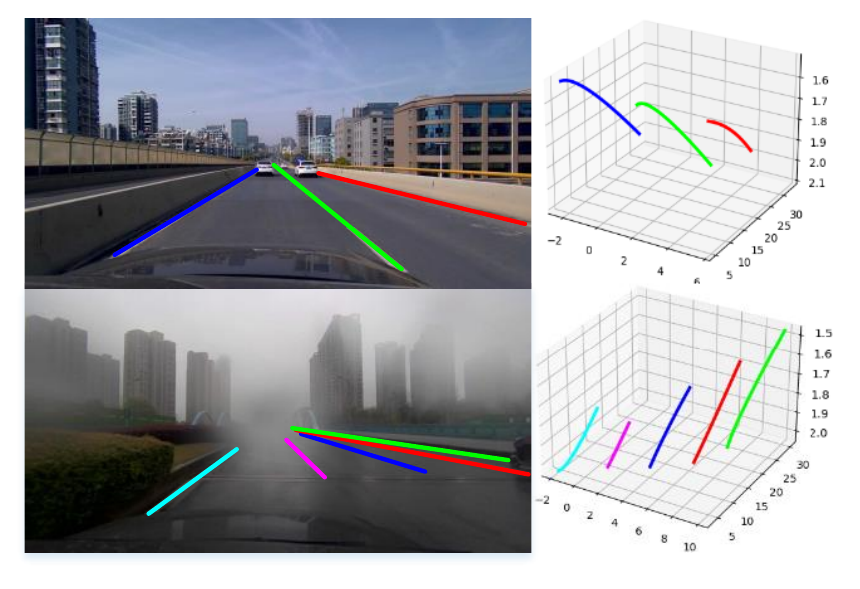}
\caption{Images and 3D lane examples of ~\textit{ONCE-3DLanes} dataset. 
\textit{ONCE-3DLanes} covers various locations, illumination conditions, weather conditions and with numerous slope scenes.}
\label{fig:examples}
\vspace{-15pt}
\end{figure}

Most existing image-based lane detection methods have exclusively focused on formulating the lane detection problem as a 2D task~\cite{2021polylanenet,2020laneatt,2018scnn}, in which a typical pipeline is to firstly detect lanes in the image plane based on semantic segmentation or coordinate regression and then project the detected lanes in top view by assuming the ground is flat~\cite{2018Lanenet,2021structureguided}.
With well-calibrated camera extrinsics, the inverse perspective mapping (IPM) is able to obtain an acceptable approximation for 3D lane in the flat ground plane.
However, in real-world driving environment, roads are not always flat~\cite{20193Dlanenet} and camera extrinsics are sensitive to vehicle body motion due to speed change or bumpy road, which will lead to the incorrect perception of the 3D road structure and thus unexpected behavior may happen to the autonomous driving vehicle.

To overcome above shortcomings associated with flat-ground assumption, 3D-LaneNet~\cite{20193Dlanenet}, 
directly predicts the 3D lane coordinates in an end-to-end manner, in which the camera extrinsics are predicted in a supervised manner to facilitate the projection from image view to top view.
In addition, an anchor-based lane prediction head is proposed to produce the final 3D lane coordinates from the virtual top view. 
Despite the promising result exhibits the feasibility of this task, the virtual IPM projection is difficult to learn without the hard-to-get extrinsics and the model is trained under the assumption that zero degree of camera roll towards the ground plane.
Once the assumption is challenged or the need of extrinsic parameters is not satisfied, this method can barely work.

In this work, we take steps towards addressing above issues.
For the first time, we present a real-world 3D lane detection dataset \textit{ONCE-3DLanes}, consisting of 211K images with labeled 3D lane points.
Compared with previous 3D lane datasets, our dataset is the largest real-world lane detection dataset published up to now, containing more complex road scenarios with various weather conditions, different lighting conditions as well as a variety of geographical locations. 
An automatic data annotation pipeline is designed to minimize the manual labeling effort.
Comparing to the method~\cite{20193Dlanenet} of using multi-sensor and expensive HD maps, ours is simpler and easier to be implemented.
In addition, we introduce a spatial-aware lane detection method, dubbed \textit{SALAD}, in an extrinsic-free and end-to-end manner.
Given a monocular input image, \textit{SALAD} directly predicts the 2D lane segmentation results and the spatial contextual information to reconstruct the 3D lanes without explicit or implicit IPM projection.

The contributions of this work are summarized as follows:
(i) For the first time, we present a largest 3D lane detection dataset \textit{ONCE-3DLanes}, alongside a more generalized evaluation metric to revive the interest of such task in a real-world scenario;
(ii) We propose a method, \textit{SALAD} that directly produce 3D lane layout from a monocular image without explicit or implicit IPM projection.

\section{Related work}
\label{sec:related work}

\subsection{2D lane detection}
There are various methods ~\cite{2020laneatt,2021laneaf,2021folo,2021condlanenet,2019SAD,2018scnn} proposed to tackle the problem of 2D lane detection.
Segmentation-based methods~\cite{2021keyseg,2018Lanenet,2021fastseg,2018scnn} predict pixel-wise segmentation labels and then cluster the pixels belonging to the same label together to predict lane instances.
Proposal-based methods~\cite{2021polylanenet,2020laneatt} first generate lane proposals either from the vanishing points~\cite{2021structureguided} or from the edges of image~\cite{2019linecnn}, and then optimize the lane shape by regressing the lane offset.
There are some other methods ~\cite{2018Lanenet,2021structureguided,2016accurate,2019project} trying to project the image into the top view and using the properties that lanes are almost parallel and can be fitted by lower order polynomial in the top view to fit lanes.
However, most methods are limited in the image view, lack the image-to-world step or suffer from the untenable flat-ground assumption.
As a result, formulating lane detection as a 2D task may cause inappropriate behaviors for autonomous vehicle when encountering hilly or slope roads~\cite{2021DualATTENTION}.

\subsection{3D lane detection}
\noindent\textbf{LiDAR-based lane detection.}
Several methods~\cite{2018laneLiDAR,2010laneLiDAR,2008lidar} have been proposed using LiDAR to detect 3D lanes,
~\cite{intensitylidar} use the characteristic that the intensity values of different material are different to filter out the point clouds of the lanes by a certain intensity threshold and then cluster them to obtain 3D lanes. 
However, it's hard to determine the specific intensity threshold since the material used in different countries or regions is distinct, and the intensity value varies much in various weather conditions, \eg, rainy or snowy.

\noindent\textbf{Multi-sensor lane detection.}
Other methods~\cite{2019lidar-camera,2021lidar-camera} try to aggregate information from both camera and LiDAR sensors to tackle lane detection task.
Specifically, ~\cite{2018Multi-Sensor} predicts the ground height from LiDAR points to project the image to the dense ground. It combines the image information with LiDAR information to produce lane boundary detection results.
Nevertheless, it's difficult to guarantee that the image and the point clouds appear in pairs in the real scenes, \eg, CULane dataset only contains images.

\noindent\textbf{Monocular lane detection.}
Recently there are a few methods~\cite{20193Dlanenet,20203Dlanenet+,2020Genlanenet,2020semi-local3dlane,2021DualATTENTION} trying to address this problem by directly predicting from a single monocular image. 
The pioneering work 3D LaneNet~\cite{20193Dlanenet} 
predicts camera extrinsics in a supervised manner to learn the inverse perspective mapping(IPM) projection, by combining the image-view features with the top-view features.
Gen-LaneNet~\cite{2020Genlanenet} proposed a new geometry-guided lane anchor in the virtual top view.
By decoupling the learning of image segmentation and 3D lane prediction, it achieves higher performance and are more generalizable to unobserved scenes.
Instead of associating each lane with a predefined anchor, 3D-LaneNet+~\cite{20203Dlanenet+} proposes an anchor-free, semi-local representation method to represent lanes.
Although the ability to detect more lane topology structures shows the anchor-free method's power.
However, all the above methods need to learn a projection matrix in a supervised way to align the image-view features with top-view features, which may cause height information loss. 
While our proposed method directly regresses the 3D coordinates in the image view without considering camera extrinsics.

\subsection{Lane datasets}
Existing 3D lane detection datasets are either unpublished or synthesized in simulated environment.
Gen-Lanenet~\cite{2020Genlanenet} uses Unity game engine to build 3D worlds and releases a synthetic 3D lane dataset, Apollo-Sim-3D, containing 10.5K images.
3D-LaneNet~\cite{20193Dlanenet} adopts a graphics engine to model terrains using a Mixture of Gaussians distribution. Lanes modeled by a $4^{th}$ degree polynomial in top view are placed on the terrains to generate the synthetic 3D lanes dataset \textit{synthetic-3D-lanes}, containing 306K images with a resolution of $360 \times 480$.
A real 3D lane dataset \textit{Real-3D-lanes} with 85K images is also created using multiple sensors including camera, LiDAR scanner and IMU as well as the expensive HD maps in~\cite{20193Dlanenet}.

In this paper, we publish the first real-world 3D lane dataset \textit{ONCE-3DLanes} which contains 211K images and covers abundant scenes with various weather conditions, different lighting conditions as well as a variety of geographical locations.
Comprehensive comparisons of 3D lane detection datasets are shown in Table~\ref{tab:dataset_compare}.

\newcommand{\tabincell}[2]{\begin{tabular}{@{}#1@{}}#2\end{tabular}}
\begin{table}[]
    \centering
    \setlength{\tabcolsep}{0.2mm}{
    \resizebox{8.5cm}{!}{
    \begin{tabular}{l|ccccl}
    \hline
    Datasets & \small images & \small real &\small published & \small \tabincell{c}{weather\\change} & \small \tabincell{c}{geographical\\locations} \\
    \hline
    synthetic-3D-lanes & 306K &  No  & Yes  & No &  -  \\
    \hline
    Apollo-Sim-3D & 10.5K  &  No & Yes & No & \tabincell{l}{highway,urban \\ residential area }\\
    \hline
    Real-3D-lanes & 85K  & Yes & No & -  &highway,rural \\
    \hline
    Ours & 211K & Yes & Yes & Yes & \tabincell{l}{highway,bridge \\ tunnel,suburb \\ downtown}\\
    \hline
    \end{tabular}}}
    \caption{Comparison of different 3D lane detection datasets. "-" means not mentioned. Ours is the first published real-world dataset covering different weather conditions and geographical locations.}
    \label{tab:dataset_compare}
    \vspace{-6pt}
\end{table}

\section{ONCE-3DLanes}

\subsection{Dataset introduction}
\noindent\textbf{Raw data.}
We construct our \textit{ONCE-3DLanes} dataset based on the most recent large-scale autonomous driving dataset ONCE (one million scenes) ~\cite{2021Once} considering its superior data quality and diversity.
ONCE contains 1 million scenes and 7 million corresponding images, the 3D scenes are recorded with 144 driving hours covering different time periods including morning, noon, afternoon and night, various weather conditions including sunny, cloudy and rainy days, as well as a variety of regions including downtown, suburbs, highway, bridges and tunnels. 
Since the camera data is captured at a speed of two frames per second and most adjacent frames are very similar, we take one frame every five frames to build our dataset to reduce data redundancy.
Also, the distortions are removed to enhance the image quality and improve the projection accuracy from LiDAR to camera. 
Thus, by downsampling the ONCE five times, our dataset contains 211k images taken by a front-facing camera.

\noindent\textbf{Lane representation.}
A lane $L^k$ in 3D space is represented by a series of points $\left\{ (x_i^k,y_i^k,z_i^k) \right\}_{i=1}^n $, which are recorded in the 3D camera coordinate system with unit meter. The camera coordinate system is placed at the optical center of the camera, with X-axis positive to the right, Y-axis downward and Z-axis forward.  

\noindent\textbf{Dataset analysis.}
\label{sec:Method}
\begin{figure}[t]
\centering 
\includegraphics[width = 8.5cm]{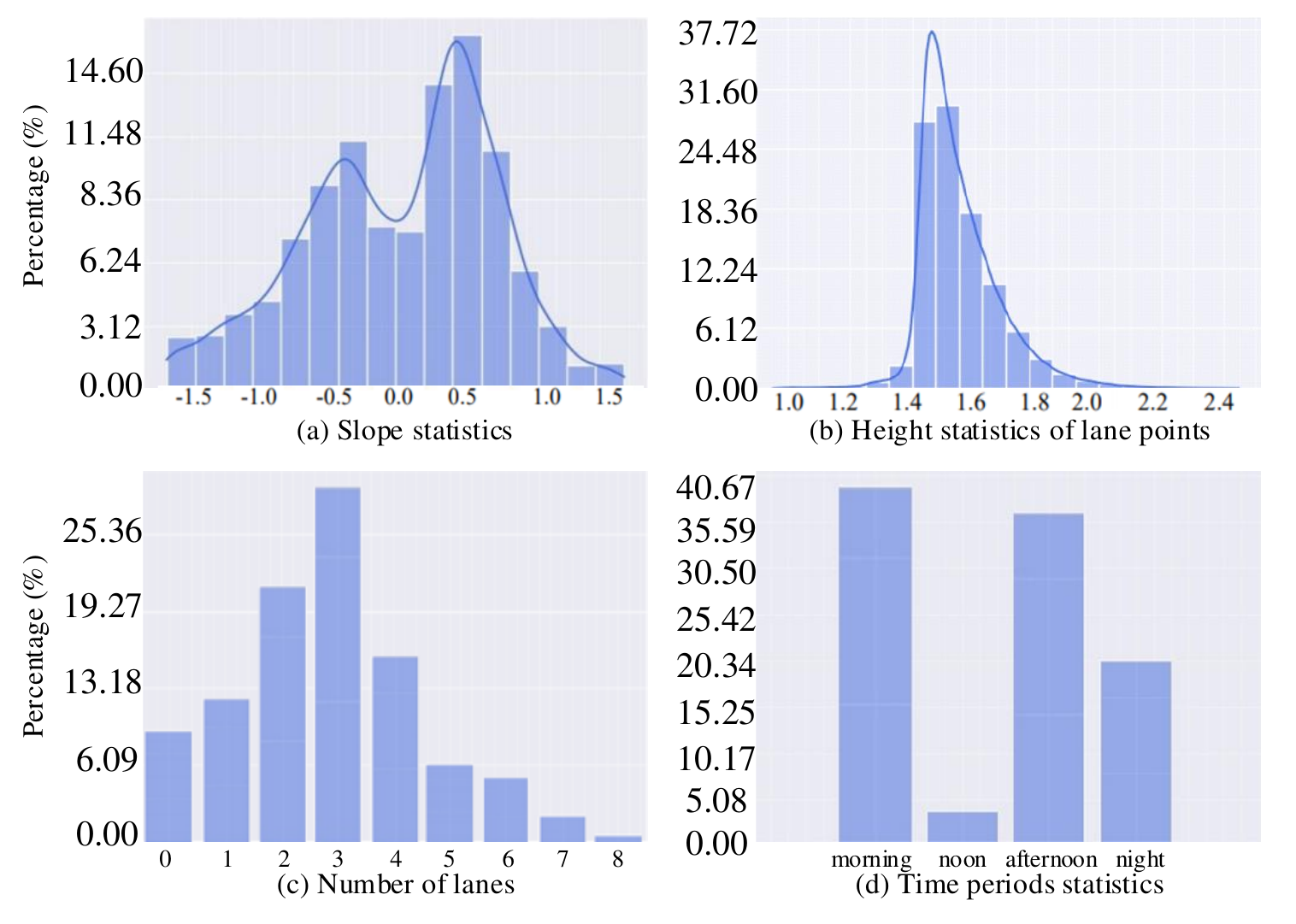}
\caption{An overview of slope scenes statistics is shown in (a). The distribution of the height of lane points is shown in (b). The histogram of the average number of lanes per image  and the time periods statistics are shown in (c) and (d) respectively.}
\label{fig:statistics}
\end{figure}
The projection error from front-view to top-view mainly occurs in the situation of slope ground, so we focus on analyzing the slope statistics on \textit{ONCE-3DLanes}.
The mean slope of lanes in each scene is utilized to represent the slope of this scene. 
The slope of a specific lane in forward direction which is considered to be the most important is calculated as follows:
\begin{equation}
    slope = (y_2 - y_1)/(z_2 - z_1)
\end{equation}
where $(x_1,y_1,z_1)$ and $(x_2,y_2,z_2)$ are the start point and the end point of the lane respectively. 
The distribution of the slope conditions and the histogram of the number of lanes per image are shown in Figure~\ref{fig:statistics}. 
It shows that our dataset is full of complexity and contains enough various slope scenes with different illumination conditions. 

\noindent\textbf{Dataset splits.}
Follow the ONCE dataset, our benchmark contains the same 3K scenes for validation and 8K scenes for testing.
To fully make use of the raw data, the training dataset not only contains the  original 5K scenes, but also the unlabeled 200K scenes. 

\begin{figure}[t]
\centering 
\includegraphics[width = 8.5cm]{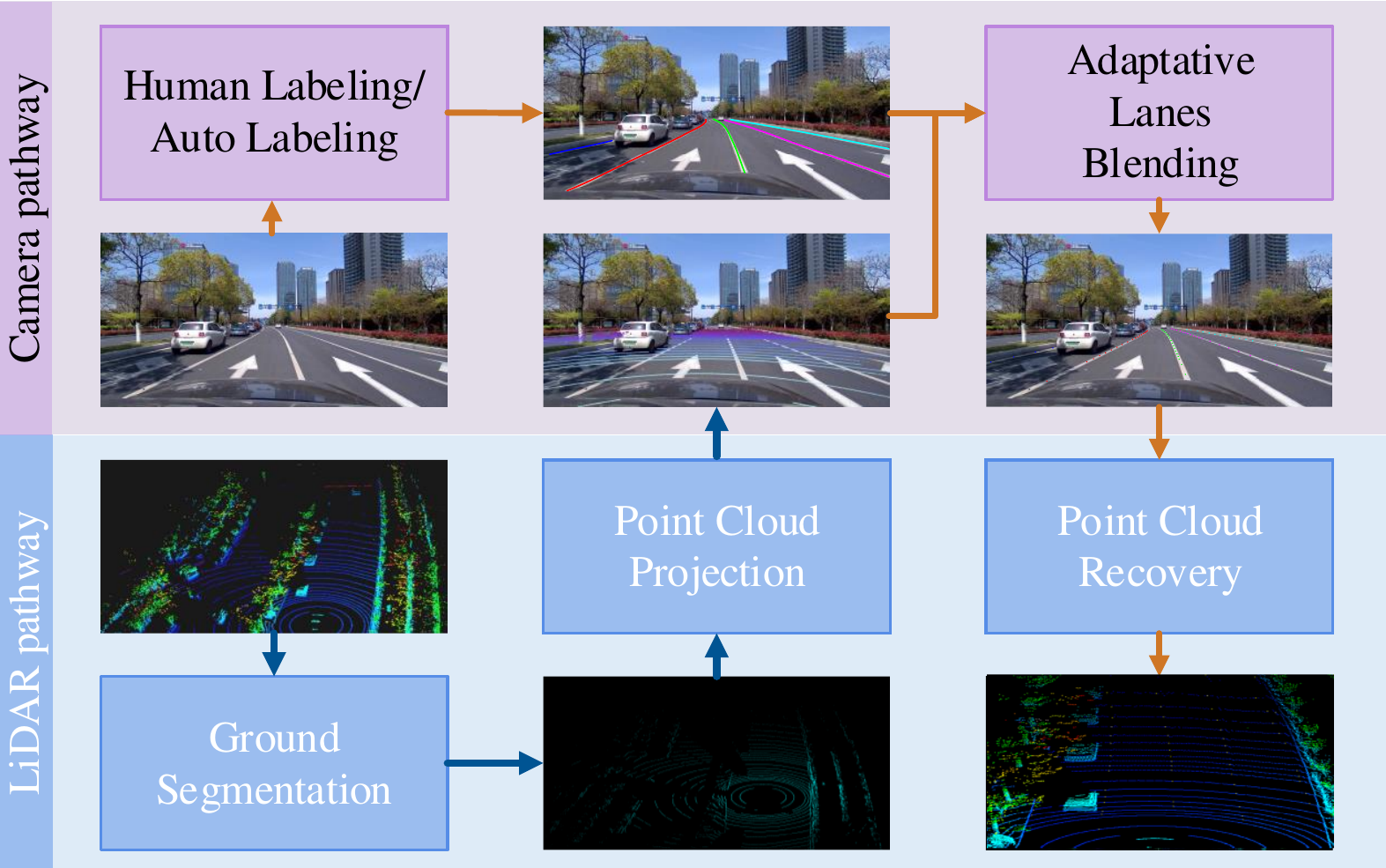}
\caption{\textbf{Dataset Annotation Pipeline}: With the paired image and Lidar point clouds as input, the 2D lanes on the image are firstly labeled and broadened to get the lane regions; secondly the ground points in the point clouds are filtered out through ground segmentation; thirdly the filtered ground points are projected to the image and collect the points which are contained in the lane regions, finally cluster the points to get the real lane points.} 
\label{fig:dataset}
\end{figure}
\subsection{Annotation pipeline}
Lanes are a series of points on the ground, which are hard to be identified in point clouds.
Hence the high-quality annotations of 3D lanes are expensive to obtain, whilst it is much cheaper to annotate the lanes in 2D images. The paired LiDAR point clouds and image pixels are thoroughly investigated and used to construct our 3D Lane dataset.
An overview of dataset construction pipeline is shown in Figure~\ref{fig:dataset}.
The pipeline consists of five steps: Ground segmentation, Point cloud projection, Human labeling/Auto labeling, Adaptive lanes blending and Point cloud recovery.
These steps are described in detail below. 

\noindent\textbf{Ground segmentation.}
Lanes are painted on the ground, which is a strong prior to locate the precise coordinate in the 3D space. To make full use of human prior and avoid the reflection of point clouds aliasing between lanes and other objects, the ground segmentation algorithm is utilized to get the ground LiDAR points at first.

The ground segmentation is performed in a coarse-to-fine manner.
In the coarse way, since the height of the LiDAR points reflected by the ground always settles in certain intervals, a pre-defined threshold is adopted to filter out those points lying on the ground coarsely based on the height statistics of the LiDAR points among the whole dataset as seen in Figure~\ref{fig:statistics}(b). In the fine way, several points in front of the vehicle are sampled randomly as seeds and then the classic region growth method is applied to get the fine segmentation result.

\noindent\textbf{Point cloud projection.}
In this step, the previous extracted ground LiDAR points are projected to the image plane with the help of calibrated LiDAR-to-camera extrinsics and camera intrinsics based on the classic homogeneous transformation, which reveals the explicit corresponding relationship between the 3D ground LiDAR points and the 2D ground pixels in the image.

\noindent\textbf{Human labeling / Auto labeling.}
To obtain 2D lane labels in the images and alleviate the taggers' burden, a robust 2D lane detector which is trained in million-level scenes is firstly used to automatically pre-annotate pseudo lane labels.
At the same time, professional taggers are required to verify and correct the pseudo labels to ensure the annotation accuracy and quality.

\noindent\textbf{Adaptative lanes blending.}
After getting the accurate 2D lane labels and ground points, in order to judge whether a ground point belongs to the lane marker or not, we broaden 2D lane labels with an appropriate and adaptive width to get lane regions. 
Due to the perspective principle, a lane is broadened with different widths according to the distance from camera.
After the point clouds projection procedure, we consider the ground points which are contained in the lane regions as the lane point clouds. 

\noindent\textbf{Point cloud recovery.}
Finally, we select these lane point clouds out. And for a specific lane, the lane point clouds in the same beam are clustered to get the lane center points to represent this lane. 

To ensure the accuracy of annotations, we do not interpolate between center points in data collection stage.
While in training stage, we use cubic spline interpolation to generate dense supervised labels.
we also compared our annotations with manually labeling results on a small portion of data, which shows the high quality of our annotations.
The interpolation code will be made public along with our dataset.

\section{SALAD}

\begin{figure*}[htp]
\centering 
\includegraphics[width = 0.95\linewidth]{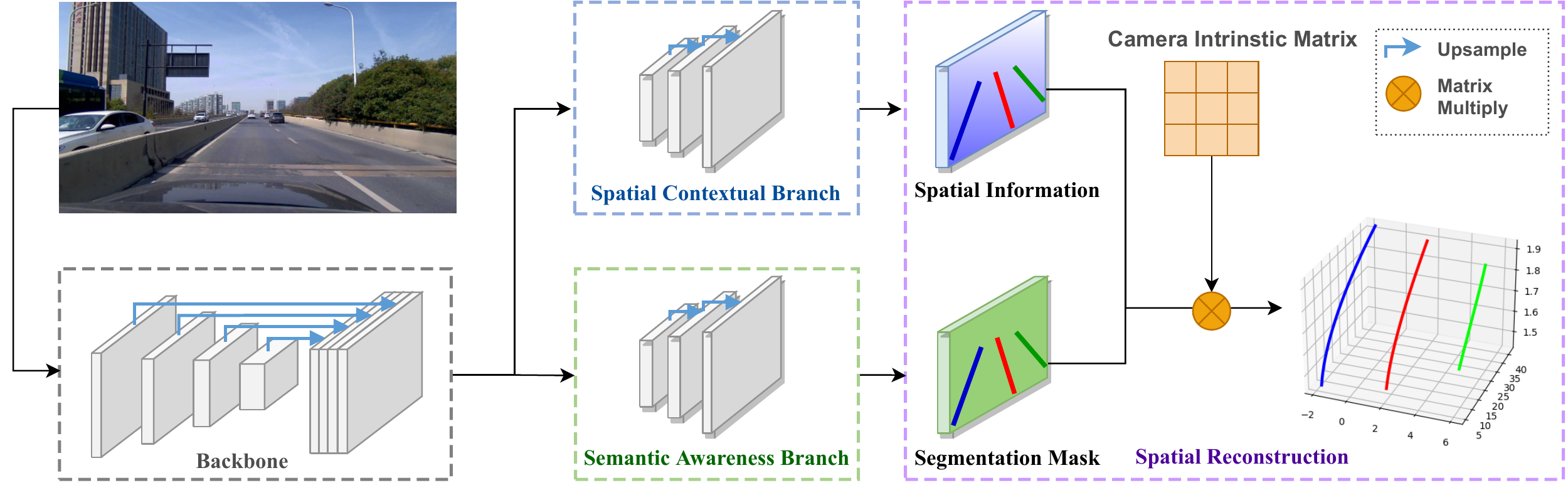}
\caption{\textbf{The architecture of \textit{SALAD}.} The backbone encodes an input image into deep features and two branches namely semantic awareness branch and spatial contextual contextual branch decode the features to get the lane's spatial information and the segmentation mask. Then 3D reconstruction is performed by integrating these information and finally obtain the 3D lane positions in real-world scene.}
\label{fig:architecture}
\end{figure*}

In this section, we introduce \textit{SALAD}, a spatial-aware monocular lane detection method to perform 3D lane detection directly on monocular images.
In contrast to previous 3D lane detection algorithms~\cite{20193Dlanenet,20203Dlanenet+,2020Genlanenet}, which project the image to top view and adopt a set of predefined anchors to regress 3D coordinates, our method does not require human-crafting anchors and the supervision of extrinsic parameters.
Inspired by SMOKE~\cite{2020smoke}, \textit{SALAD} consists of two branches: semantic awareness branch and spatial contextual branch.
The overall structure of our model is illustrated in Figure~\ref{fig:architecture}.
In addition, we also adopt a revised joint 3D lane augmentation strategy to improve the generalization ability.
The details of our network architecture and augmentation methods are discussed in the following parts.

\subsection{Backbone}
We choose the Segformer\cite{2021segformer} as our backbone to extract the global contextual features and to learn the slender structure of lanes.
To be concrete, given an image $I \in \mathbb{R}^{H \times W \times 3} $, the hierarchical transformer backbone encodes the image $I$ into multi-level features at $\left\{1/4,1/8,1/16,1/32\right\}$ of the input resolution.
Then all multi-level features are upsampled to $\frac{H}{4} \times \frac{W}{4}$ and concatenated through a MLP-based decoder and a convolutional feature fusion layer to aggregate the multi-level features into $F \in \mathbb{R}^{\frac{H}{4} \times \frac{W}{4} \times C}$.
Specifically, we adopt Segformer-B2 as our feature extractor.

\subsection{Semantic awareness branch}
Traditional 3D lane detection methods~\cite{20193Dlanenet, 20203Dlanenet+, 2020Genlanenet} directly projecting feature map from front view to top view, are not reasonable and harmful to the performance of prediction as the feature map may not be organized following the perspective principle.

In order to directly regress 3D lane coordinates, we first design a semantic awareness branch to make full use of 2D semantic information, which provides 2D lane point proposals to aggregate 3D information together.
It is a relatively easy task to extract 2D features of lane markers from images with rich semantic information.
In addition, current mature experience on semantic segmentation task can also be used to enhance our semantic awareness branch.
We follow the method of \cite{2018Lanenet}, to encode 2D lane coordinates $\left\{ (u_i^k,v_i^k) \right\}_{i=1}^n $ into ground-truth segmentation map $S_{gt} \in \mathbb{R}^{H \times W \times 1}$.

During training time, image $I \in \mathbb{R}^{H \times W \times 3}$ and ground-truth $S_{gt} \in \mathbb{R}^{H \times W \times 1}$ pairs are utilized to train semantic awareness branch.
During inference, given an image $I \in \mathbb{R}^{H \times W \times 3}$, we are able to locate the foreground lane points on the binary mask $S \in \mathbb{R}^{H \times W \times 1}$.

According to \cite{debevec1996modeling}, inverse projection from 2D image to 3D space is an underdetermined problem.
Based on the segmentation map generated by semantic awareness branch, spatial information of each pixel is also required to transfer this segmentation map from 2D image plane to 3D space.

\subsection{Spatial contextual branch}
To reconstruct 3D lanes from 2D lane points generated by semantic awareness branch, we propose the spatial contextual branch to predict vital 3D offsets.
To sum up, our spatial contextual branch predicts a regression result $O = [\delta_{u}, \delta_{v}, \delta_{z}]^{T} \in \mathbb{R}^{3 \times H \times W}$.
$\delta_{u}$ and $\delta_{v}$ denote the pixel location offsets of lane points predicted in segmentation branch $(u_{s}, v_{s})$, and generate accurate 2D lane positions.
The $\delta_{z}$ denotes the pixel-level prediction of depth.

Due to the downsampling and lacking of global information, the locations of the predicted lane points are not accurate enough.
Our spatial contextual branch accepts feature $F$ and outputs a pixel-level offset map, which predicts the spatial location shift $\delta_u$ and $\delta_v$ of lane points along $u$ and $v$ axis on the image plane.
With the predictions of pixel location offsets $\delta_u$ and $\delta_v$, the rough estimation of lane point locations is modified by global spatial context:

\begin{equation}
    \begin{bmatrix} u \\ v \end{bmatrix} = \begin{bmatrix}
    u_s + \delta_{u} \\ v_s + \delta_{v}
    \end{bmatrix}.
\end{equation}

In order to recover 3D lane information, the spatial contextual branch also generates a dense depth map to regress on depth offset $\delta_z$ for each pixel of the lane markers.
Considering the depth of the ground on image plane increases along rows, we assign each row of the depth map a predefined shift $\alpha_r$ and scale $\beta_r$, and perform regression in a residual way.
The standard depth value $z$ is recovered as following:

\begin{equation}
    z = \alpha_{r} + \beta_{r}\delta_{z}.
\end{equation}

The ground-truth depth map is generated by projecting 3D lane points $\left\{ (x_i^k,y_i^k,z_i^k) \right\}_{i=1}^n $ on the image plane to get pixel coordinates $\left\{ (u_i^k,v_i^k, z_i^k) \right\}_{i=1}^n $.
Then at each pixel $(u_i^k, v_i^k)$, its corresponding depth value is assigned to $z_i^k$.
Following ~\cite{ku2018defense}, we apply depth completion on the sparse depth map to get the dense depth map $D_{gt}$ to provide sufficient training signals for our spatial contextual branch.

\subsection{Spatial reconstruction}
The spatial information predicted by our spatial contextual branch of our model plays a virtual role in 3D lane reconstruction.
To map 2D lane coordinates back to 3D spatial location in camera coordinate system, the depth information is an indispensable element.
To be concrete, given the camera intrinsic matrix $K_{3\times3}$, 
a 3D point $(x,y,z)$ in camera coordinate system can be projected to a 2D image pixel $(u,v)$ as:
\begin{equation}
z\begin{bmatrix} u \\ v \\1 \end{bmatrix} = K_{3\times3} \begin{bmatrix} x \\ y \\z \end{bmatrix} =  \begin{pmatrix} 
f_x & s & c_x \\
0 & f_y & c_y \\
0 & 0 & 1 
\end{pmatrix}
\begin{bmatrix} x \\ y \\z \end{bmatrix},
\end{equation}
where $f_x$ and $f_y$ represent the focal length of the camera, $(c_x,c_y)$ is the principal point and $s$ is the axis skew.
Thus, given a 2D lane point in the image with pixel coordinates $(u,v)$ along with its depth information $d$, noted that the depth denotes the distance to the camera plane, so the depth $d$ is the same as the $z$ in the camera coordinate system.    
Thus 3D lane point in camera coordinate system $(x, y, z)$ can be restored as follows:
\begin{equation}
\left\{
\begin{aligned}
z & = d, \\
y & = \frac{z}{f_y} \cdot (v-c_y), \\
x & = \frac{z}{f_x} \cdot [(u-c_x) - \frac{s}{f_y}(v-c_y)].
\end{aligned}
\right.
\end{equation}

Utilizing the fixed parameters of the camera intrinsic, we can project the 2D lane proposal points back to 3D locations to reconstruct our 3D lanes.

\subsection{Loss function}
Given an image and its corresponding ground-truth 3D lanes, the loss function between predicted lanes and ground-truth lanes are formulated as:
\begin{equation}
    \mathcal{L} = \mathcal{L}_{seg} + \lambda \mathcal{L}_{reg}.
\end{equation}
$\mathcal{L}_{seg}$ is for the binary segmentation branch, with a cross-entropy loss in a pixel-wise manner on the segmentation map.
For a specific pixel in segmentation map $S$, $y_i$ is the label and $p_i$ is the probability of being foreground pixels:
\begin{equation}
    \mathcal{L}_{seg} = - \frac{1}{N}\sum_{i=1}^{N}[y_ilog(p_i) + (1-y_i)log(1-p_i)].
\end{equation}
$\mathcal{L}_{reg}$ is for the spatial contextual branch, which predicts the spatial offsets $O = [\delta_{u}, \delta_{v}, \delta_{z}]^{T}$.
we choose smooth L1 loss to regress these spatial contextual information $O$:
\begin{equation}
\begin{aligned}
\mathcal{L}_{reg} = &\frac{1}{N}\sum_{i=1}^{N}[smooth_{L_1}(\hat{O}_i - O_i)].
\end{aligned}
\end{equation}
$\lambda$ denotes the penalty term for regression loss and is set to $1$ in our experiments.

\subsection{Data augmentation}
Randomly horizontal flip and image scaling are common data augmentation methods to improve the generalization ability of 2D lane detection models.
However, it is worth noting that the image shift and scale augmentation methods will cause 3D information inconsistent with the data augmentation~\cite{2020smoke}.
We revise it by proposing a joint scale strategy. 

To ensure we can restore the same size of the original image from the scaled image, we first crop the top of the image with size c. 
Then we scale the cropped image with the proportion s.
According to the similar triangles theorems, it is proved that the relationship of 3D information of a specific pixel before scaling $(x,y,z)$ and after scaling $(\hat{x},\hat{y},\hat{z})$ is:
\begin{equation}
    (\hat{x},\hat{y},\hat{z}) = (x,y,z \cdot s). 
\end{equation}
Take scaling factor $s$ which is less than one for example. 
If the image is scaled with factor $s$, in the camera coordinate system, it is like the camera is moving forward in the $Z$ direction.
As for a specific point, the $x$ and $y$ keep the same while the $z$ become smaller and the new $z$ equals $z \cdot s$.
Using this strategy, we can ensure that the 3D ground truth remain consistent during 2D image data augmentation.

\section{Experiment}
\label{sec:formatting}

In this section, our experiments are presented as follows.
First we introduce our experimental setups, including evaluation metrics and implementation details.
Then we evaluate our baseline method on our \textit{ONCE-3DLanes} dataset and investigate the evaluation performance of different hyper-parameter settings.
Next we compare our proposed method with the prior state-of-the-art to prove the superiority of our proposed method.
Finally, we conduct several ablations studies to show the significance of modules in our network.

\subsection{Evaluation metric}
Evaluation metric is set to measure the similarity between the predicted lanes and the ground-truth lanes.
Previous evaluation metric~\cite{2020Genlanenet}, which set predefined $y$-position to regress the $x$ and $z$ coordinates of lane points, is not sophisticated. 
Due to the fixed anchor design, this metric essentially performs badly when the lanes are horizontal.

To tackle this problem, we propose a two-stage evaluation metric, which regards 3D lane evaluation problem as a point cloud matching problem combined with top-view constraints.
Intuitively, two 3D lane pairs are well matched when achieving little difference in the $z$-$x$ plane (top view) alongside a close height distribution.
Matching in the top view constrains the predicted lanes in the correct forward direction, and close point clouds distance in 3D space ensures the accuracy of the predicted lanes in spatial height.

\begin{figure}[htp]
\centering 
\includegraphics[width = 8cm]{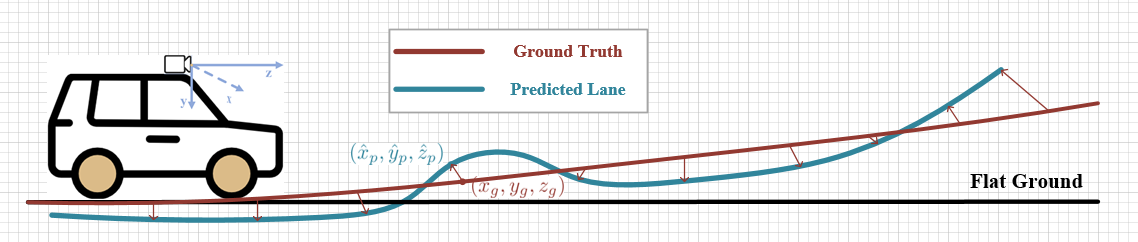}
\caption{\textbf{Unilateral chamfer distance} Given a point on the gound-truth lane, find the nearest point on the predicted lane to calculate the chamfer distance.}
\label{fig:CD}
\end{figure}

Our proposed metric first calculate the matching degree of two lanes on the $z$-$x$ plane. 
To be concrete, lane is represented as $L^k = \left\{ (x_i^k,y_i^k,z_i^k) \right\}_{i=1}^n $.
To judge whether predicted lane $L^p$ matches ground-truth lane $L^g$, the first matching process is done in the z-x plane, namely top-view, 
we use the traditional IoU method~\cite{2018scnn} to judge whether $L^p$ matches $L^g$. 
If the IoU is bigger than the IoU threshold, further we use a unilateral Chamfer Distance ($CD$) to calculate the curves matching error in the camera coordinates.
The curve matching error $CD_{p,g}$ between $L^p$ and $L^g$ is calculated as follows:
\begin{small}
\begin{equation}
\left\{
\begin{aligned}
& CD_{p,g} = \frac{1}{m} \sum_{i=1}^{m} || P_{g_i} - \hat{P}_{p_j}||_2, \\
&\hat{P}_{p_j} = \mathop{min}\limits_{P_{p_j}\in L^p}||P_{p_j}-P_{g_i}||_2, \\
\end{aligned}
\right.
\end{equation}
\end{small}

where $P_{p_j}=(x_{p_j}, y_{p_j}, z_{p_j})$ and $P_{g_i}=(x_{g_i}, y_{g_i}, z_{g_i})$ are point of $L^p$ and $L^g$ respectively, and $\hat{P}_{p_j}$ is the nearest point to the specific point $P_{g_i}$.
$m$ represents the number of points token at an equal distance from the ground-truth lane. 
If the unilateral chamfer distance is less than the chamfer distance threshold, written as $\tau_{CD}$.
We consider $L^p$ matches $L^g$ and accept $L^p$ as a true positive. 
The legend to calculate chamfer distance error is shown in Figure~\ref{fig:CD}.

This evaluation metric is intuitive and strict, and more importantly,
it applies to more lane topologies such as vertical lanes, so it is more generalized.
At last, since we know how to judge a predicted lane is true positive or not, we use the precision, recall and F-score as the evaluation metric.

\subsection{Implementation details} 

Our experiments are carried out on our proposed \textit{ONCE-3DLanes} benchmark.
We use the Segformer~\cite{2021segformer} as our backbone with two branches.
The Segformer encoder is pre-trained with Imagenet~\cite{2012imagenet}.
The input resolution is set to 320 $\times$ 800 with our augmentation strategy during training.
The augmentation is turned off during testing.
The Adamw optimizer is adopted to train 20 epochs, with an initial learning rate at 3e-3 and applied with a poly scheduler by default.
For evaluation, We set the IoU threshold as 0.3, the chamfer distance thresh $\tau_{CD}$ as 0.3m.
We also test our model using the Mindspore~\cite{mindspore}.

\subsection{Benchmark performance}
\subsubsection{Main results}

We train our model with the whole training set of 200k images and report the detection performance on the test set of \textit{ONCE-3DLanes} dataset.
To verify the rationality of the hyper-parameters setting in our evaluation metric, we evaluate our model under different Chamfer Distance threshold $\tau_{CD}$ and report the testing results in Table~\ref{tab:main result}.

\begin{table}[htb]
	\centering
	\setlength{\tabcolsep}{0.7mm}{
	\begin{tabular}{l|cccc}
		$\tau_{CD}$(m)& F1(\%) & Precision(\%) & Recall(\%) & CD error(m)\\ 
		\hline
		0.15 & 48.35 & 58.52 & 41.19 & 0.062\\
		0.30 & 64.07 & 75.90 &55.42 & 0.098 \\
		0.50 & 68.92 & 81.65 & 59.62 & 0.118 \\
	\end{tabular}}
	\vspace{-5pt}
	\caption{Performance of \textit{SALAD} under different $\tau_{CD}$ thresholds on our test set.}
	\label{tab:main result}
	\vspace{-5pt}
\end{table}

\begin{figure}[!t]
    \centering
    \includegraphics[width = 8.5cm]{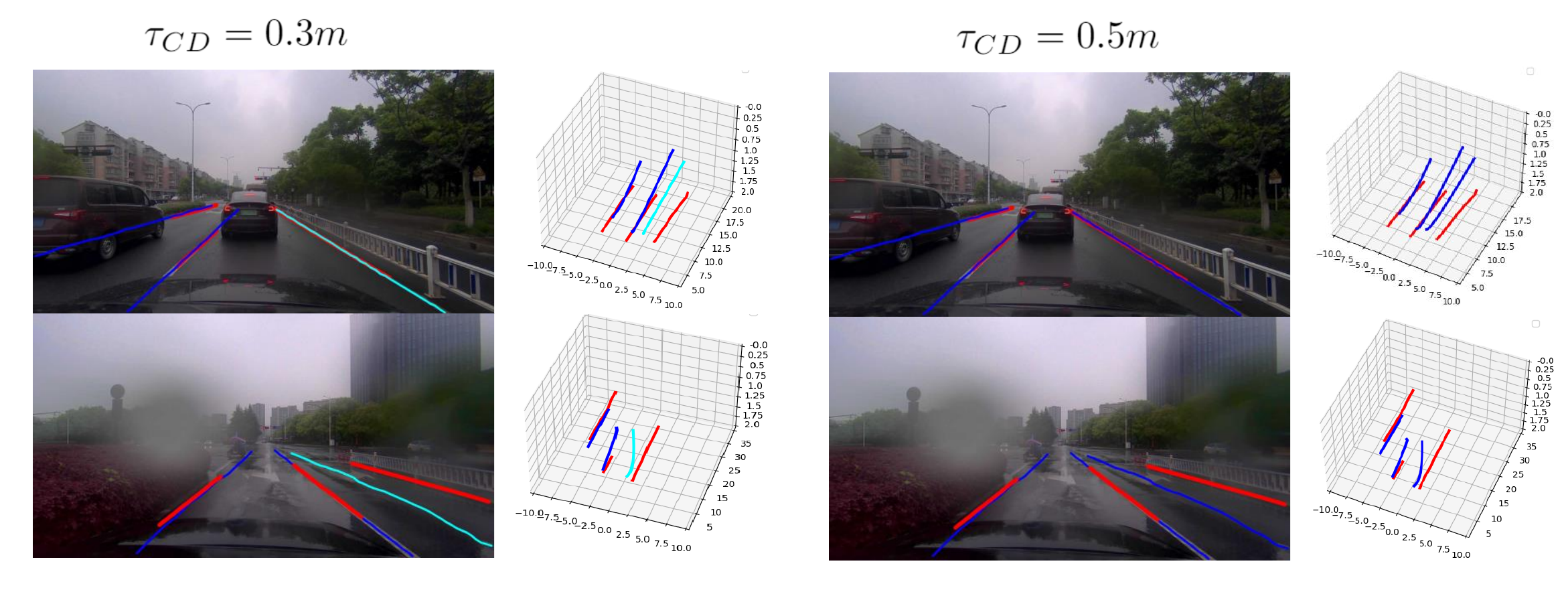}
    \caption{Qualitative results and failure cases analysis under specific threshold. The ground-truth lanes are colored in red, while true positives of our predictions are in blue and false positives in cyan. The $\tau_{CD}$ of 0.5 is somehow too loose for discrimination.}
    \label{fig:case}
    \vspace{-5pt}
\end{figure}

We report our performance in different settings of $\tau_{CD}$, in order to fully investigate the impact of tightening or loosening the criteria on model performance.
The illustrations of criteria adopting various $\tau_{CD}$ are also proposed in Figure~\ref{fig:case}.
It can be seen that under the threshold of 0.5, some predicted lanes relatively far from the ground-truth are judged as true positives.
While under the $\tau_{CD}$ of 0.15, the criteria seems too harsh.
The $\tau_{CD}$ of 0.3 is more reasonable and our \textit{SALAD} achieves $64.07\%$ F1-score based on this threshold.
Besides, since the distance is calculated based on real scene so it is highly adaptable to real-world datasets.
In the remaining parts, the experimental results are reported under the threshold of 0.3.

\subsubsection{Results of 3D lane detection methods}
In order to further verify the authenticity of our dataset and the superiority of our method, which is extrinsic-free, we also conduct some experiments of other 3D lane detection algorithms on our dataset.

It is worth noting that all existing 3D lane detection algorithms need to provide camera poses as supervision information and have strict assumptions that the camera is installed at zero degrees roll relative to the ground plane~\cite{20193Dlanenet}.
However, our method requires no external parameter information.
To make comparison, we used the camera-pose parameters provided by ONCE~\cite{2021Once} to provide supervision signals for counterpart methods and ultimately evaluated them on our test set.
The performance of 3D lane detection algorithms are presented in Table ~\ref{tab:three-d result}.

\begin{table}[t]
	\centering
	\scalebox{0.80}{
	\setlength{\tabcolsep}{0.5mm}{
	\begin{tabular}{l|ccccc}
		Method & \textit{extri. free} & F1(\%) & Precision(\%) & Recall(\%) & CD error(m)\\ 
		\hline
		3D LaneNet~\cite{20193Dlanenet} & \xmark & 44.73 & 61.46 & 35.16 & 0.127\\
		GenLaneNet~\cite{2020Genlanenet} & \xmark & 45.59 & 63.95 & 35.42 & 0.121 \\
		\hline
		\textit{SALAD} & \cmark & \textbf{64.07} & 75.90 &55.42 & 0.098\\
	\end{tabular}}}
	\vspace{-5pt}
	\caption{Performance of 3D lane detection on ONCE-3DLanes.}
	\label{tab:three-d result}
	\vspace{3pt}
\end{table}

The comparison shows our method outperforms other 3D lane detection method in \textit{ONCE-3DLanes} dataset.
The comparison result indicates that extrinsic-required methods under the assumption of fixed camera pose and zero-degree camera roll may suffer in real 3D scenarios.

\subsubsection{Results of extended 2D lane detection methods}

\textit{ONCE-3DLanes} dataset is a newly released 3D lane detection dataset and no previous work addresses the problem of extrinsic-free 3D lane detection.
In order to verify the validity of our dataset and the efficiency of our method, we extend the existing 2D lane detection model and evaluate their performance on our dataset.

Different from 3D lane detection methods, 2D lane detection algorithms can only detect the pixel coordinates of lanes in the image plane, but cannot recover the spatial information of lanes.
To obtain 3D lane detection results, we use a pre-trained depth estimation model MonoDepth2~\cite{2019monodepth2} (finetuned to the depth scale of our dataset) to estimate the pixel-level depth of the image.
It is worth mentioning that the depth model is finetuned on full ONCE dataset in order to avoid under-fitting caused by sparse supervision provided by lane points, which also indicates that this pipeline is difficult to be extended on other 3D lane benchmarks.
Combined with the detection results of extended 2D lane detection model, the spatial position of 3D lanes are reconstructed and our evaluation metric is used for performance evaluation.
The results are showed in Table \ref{tab:two-d result}.

\begin{table}[htb]
	\centering
	\scalebox{0.85}{
	\setlength{\tabcolsep}{1mm}{
	\begin{tabular}{l|cccc}
		Method  & F1(\%) & Precision(\%) & Recall(\%) & CD error(m)\\ 
		\hline
		\small PointLaneNet (R101)~\cite{2019pointlanenet} & 54.99 & 64.50 & 47.93 & 0.115\\
		UltraFast (R101)~\cite{2020ultrafast} & 54.18 & 63.68 & 47.14 & 0.128 \\
		RESA (R101)~\cite{2021resa} & 55.53 & 65.08 & 48.43 & 0.112 \\
		LaneAF (DLA34)~\cite{2021laneaf} & 56.39 & 66.07 & 49.18 & 0.109 \\
		LaneATT (R122)~\cite{2020laneatt} & 56.57 & 66.75 & 49.07 & 0.101 \\
		\hline
		\textit{SALAD} & \textbf{64.07} & 75.90 & 55.42 & 0.098 \\
	\end{tabular}}}
	\vspace{5pt}
	\caption{Performance of extended 2D lane detection methods on \textit{ONCE-3DLanes} test set.}
	\label{tab:two-d result}
	\vspace{3pt}
\end{table}

Experimental results show that the extended 2D models are effective to perform 3D lane detection task on our~\textit{ONCE-3DLanes} dataset.
It can also be found that our proposed method can reach $64.07\%$ F1-score, outperforms the best of other methods $56.57\%$ by $7.5\%$, which shows the superiority of our method.

\subsection{Ablation study}

To verify the efficiency of revised data augmentation methods, we conduct ablation experiments by gradually turning off data augmentation strategy.
As shown in Table~\ref{tab:dataaug}, the performance of our method is constantly improved with the gradual introduction of our data augmentation strategy.
The flip method brings a $1.06\%$ improvement to our model and the 3D scale provides a further $1.83\%$ improvement, which proves the validity of our augmentation strategy.

\begin{table}[htb]
	\centering
	\scalebox{0.88}{
	\setlength{\tabcolsep}{0.5mm}{
	\begin{tabular}{l|cccc}
		Method  & F1(\%) & Precision(\%) & Recall(\%) & CD error(m)\\ 
		\hline
		\textit{SALAD} (w/o aug)  & 61.18 & 72.43 & 52.95 & 0.103\\
		\hline
	    + flip & $62.24^{\textbf{+1.06}}$ & $73.67^{\textbf{+1.24}}$ & $53.88^{\textbf{+0.93}}$ & $0.102^{\textbf{-0.001}}$ \\
	    + joint scale & $64.07^{\textbf{+2.89}}$ & $75.90^{\textbf{+3.47}}$ & $55.42^{\textbf{+2.47}}$ & $0.098^{\textbf{-0.005}}$ \\
	\end{tabular}}}
	\vspace{-5pt}
	\caption{Ablation study on data augmentation strategy.} 
	\label{tab:dataaug}
	\vspace{-6pt}
\end{table}

\section{Conclusion and limitations}
\label{sec:conclusion}

In this paper, we have presented a largest real-world 3D lane detection benchmark \textit{ONCE-3DLanes}.
To revive the interest of 3D lane detection, we benchmark the dataset with a novel evaluation metric and propose an extrinsic-free and anchor-free method, dubbed \textit{SALAD}, directly predicting the 3D lane from a single image in an end-to-end manner.
We believe our work can lead to the expected and unexpected innovations in communities of both academia and industry.

As our dataset construction requires LiDAR to provide 3D information,
occlusions would lead to short interruption and we have used interpolation to fix it.
Missing points problem in the distance still exists due to the low resolution of LiDAR.
Future work will focus on the ground point clouds completion to generate full information for 3D lanes.

\paragraph{Acknowledgments}
This work was supported in part by 
National Natural Science Foundation of China (Grant No. 6210020439),
Lingang Laboratory (Grant No. LG-QS-202202-07),
Natural Science Foundation of Shanghai (Grant No. 22ZR1407500),
Shanghai Municipal Science and Technology Major Project (Grant No. 2018SHZDZX01 and 2021SHZDZX0103),
Science and Technology Innovation 2030 - Brain Science and Brain-Inspired Intelligence Project (Grant No. 2021ZD0200204),
MindSpore and CAAI-Huawei MindSpore Open Fund.

{\small
\bibliographystyle{ieee_fullname}
\normalem
\bibliography{egbib}
}

\appendix

\section{Appendix}

\subsection{Further analysis}
We conduct ablation studies to show the rationality of our experiment settings including loss function,  backbone network and regression method.

\paragraph{Loss function}
As shown in Table~\ref{tab:loss}, for the spatial contextual branch, we study different loss functions. Results show the smooth L1 loss outperforms L1 and L2 loss functions at all metrics.

\paragraph{Backbone network}
We compare the SegFormer~\cite{2021segformer} with Unet~\cite{ronneberger2015u} for the backbone network.
Moreover, different attention mechanisms~\cite{Vaswani2017AttentionIA,wang2020axial} are added to Unet to help learn the global information of lane structures. 
Table~\ref{tab:backbone} shows that model with SegFormer beat the variants of Unet by a clear margin.

\paragraph{Regression method}
We also conduct an ablation study to evaluate the way to predict the depth information in Table~\ref{tab:depth offset}.
Our method regress in a residual manner is referred as \textit{relative} method, and the method directly regress the depth information without pre-defined shift and scale is called \textit{absolute} method. 
Table~\ref{tab:depth offset} shows the \textit{relative} outperforms \textit{absolute} by a large margin.

\begin{table}[htb]
	\centering
	\scalebox{0.85}{
	\setlength{\tabcolsep}{1mm}{
	\begin{tabular}{l|cccc}
		Loss function  & F1(\%) & Precision(\%) & Recall(\%) & CD error(m)\\ 
		\hline
		L1 & 63.47 & 75.08 & 54.97 & 0.101 \\
		L2 & 62.91 & 74.55 & 54.41 & 0.103\\
		Smooth L1 & \textbf{64.07} & \textbf{75.90} & \textbf{55.42} & \textbf{0.098} \\
	\end{tabular}}}
	\caption{Ablation studies on loss functions in the spatial contextual branch.}
	\label{tab:loss}
\end{table}
\begin{table}[htb]
	\centering
	\scalebox{0.80}{
	\setlength{\tabcolsep}{1mm}{
	\begin{tabular}{l|cccc}
		Backbone  & F1(\%) & Precision(\%) & Recall(\%) & CD error(m)\\ 
		\hline
		 Unet~\cite{ronneberger2015u} & 61.12 & 73.47 & 52.32 & 0.105\\
	     Unet+self-att.~\cite{Vaswani2017AttentionIA} & 62.71 & 74.41 & 54.19 & 0.101 \\
	     Unet+axial-att.~\cite{wang2020axial} & 63.15 & 74.81 & 54.64 & 0.101 \\
		 SegFormer-B2~\cite{2021segformer} & \textbf{64.07} & \textbf{75.90} & \textbf{55.42} & \textbf{0.098} \\
	\end{tabular}}}
	\caption{Ablation studies on backbone networks.}
	\label{tab:backbone}
\end{table}
\begin{table}[htb]
	\centering
	\scalebox{0.85}{
	\setlength{\tabcolsep}{1mm}{
	\begin{tabular}{l|cccc}
		Offset option  & F1(\%) & Precision(\%) & Recall(\%) & CD error(m)\\ 
		\hline
		\textit{absolute} & 62.37 & 74.07 & 53.86 & 0.104\\
		\textit{relative} & \textbf{64.07} & \textbf{75.90} & \textbf{55.42} & \textbf{0.098} \\
	\end{tabular}}}
	\caption{Ablation studies on depth regression methods.}
	\label{tab:depth offset}
\end{table}

\subsection{More qualitative results}
\begin{figure*}[htp]
\centering 
\includegraphics[width = 0.85\linewidth]{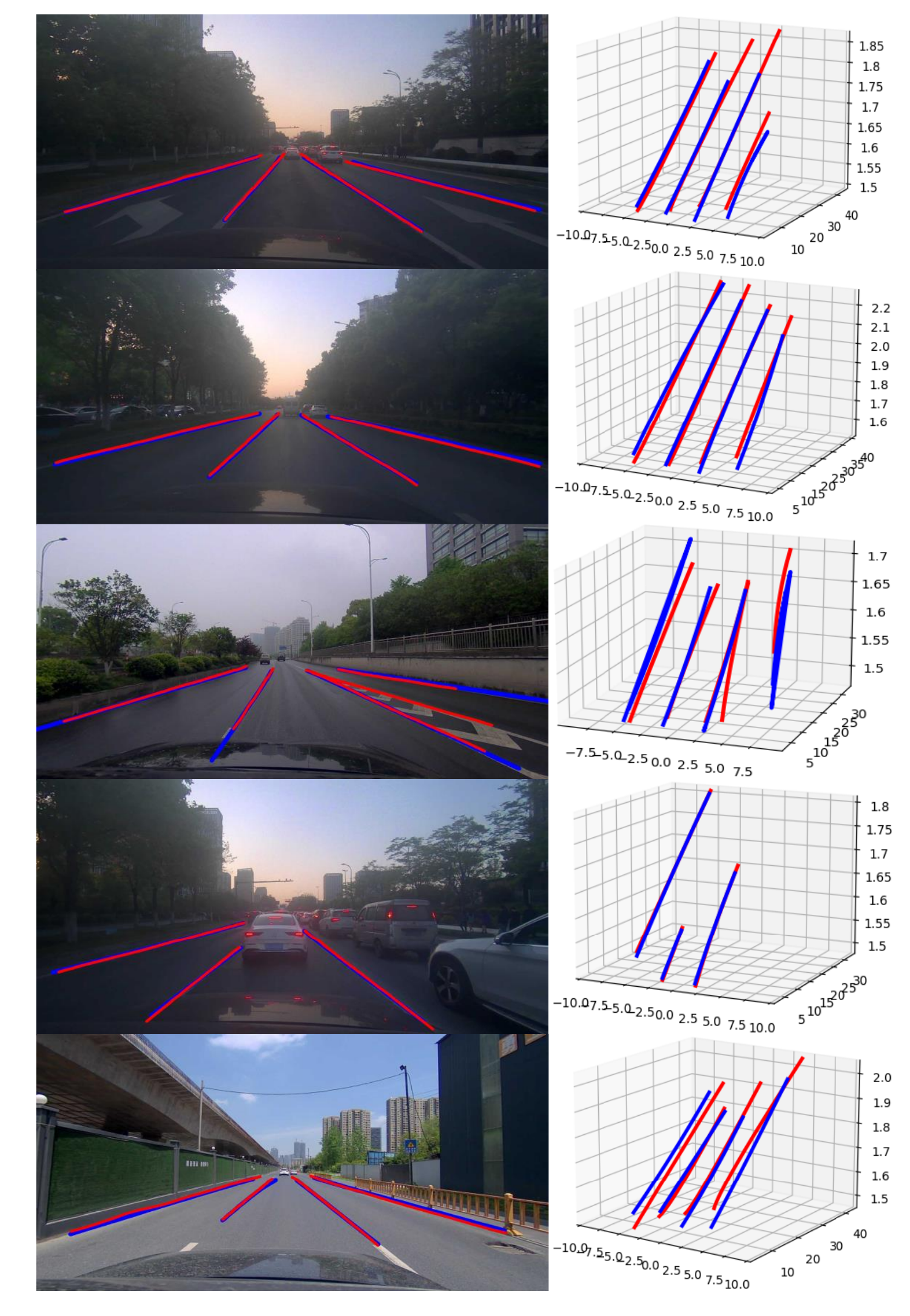}
\caption{Visualization of \textit{SALAD} on ONCE-3DLanes test set.
The ground-truth lanes are colored in red while the predicted lanes are colored in blue. 
2D projections are shown in the left and 3D visualizations in the right. }
\label{fig:ex-salad}
\end{figure*}

We present the qualitative results of \textit{SALAD} lane prediction in Figure~\ref{fig:ex-salad}.
2D projections are shown in the left and 3D visualizations are presented in the right.

\begin{figure*}[htp]
\centering 
\includegraphics[width = 0.85\linewidth]{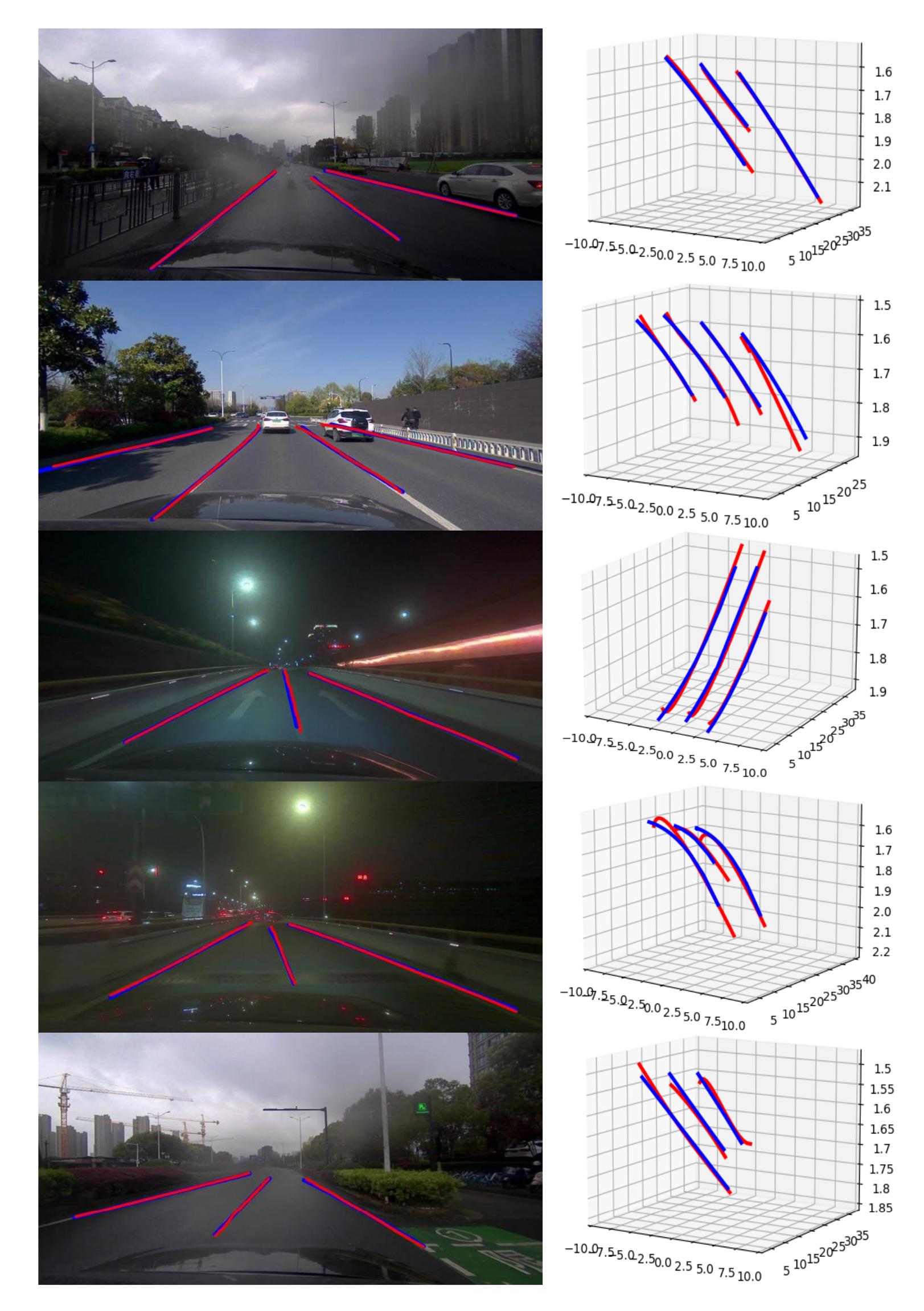}
\label{fig:results}
\end{figure*}

\end{document}